\documentclass[11pt]{article}
\usepackage[a4paper,margin=1in]{geometry}
\usepackage[T1]{fontenc}
\usepackage[latin1]{inputenc}
\usepackage{lmodern}

\usepackage{graphicx}
\usepackage{subcaption}
\usepackage{amssymb,amsmath,array}
\usepackage{booktabs}
\usepackage{pdflscape}
\usepackage{multirow}
\usepackage{pgfplots}
\pgfplotsset{compat=1.18}

\usepackage{algorithm}
\usepackage{algpseudocode}

\usepackage{tikz}
\usetikzlibrary{arrows.meta,positioning,calc}

\begin{document}

\title{Evaluating Counterfactual Explanation Methods on Incomplete Inputs}

\author{
Francesco Leofante\\
Department of Computing, Imperial College London, UK\\
\texttt{f.leofante@imperial.ac.uk}
\and
Daniel Neider\\
TU Dortmund University, Dortmund, Germany\\
\texttt{daniel.neider@tu-dortmund.de}
\and
Mustafa Yal\c{c}{\i}ner\thanks{Corresponding author.}\\
TU Dortmund University, Dortmund, Germany\\
Center for Trustworthy Data Science and Security,\\
University Alliance Ruhr, Dortmund, Germany\\
\texttt{mustafa.yalciner@tu-dortmund.de}
}

\date{}

\maketitle
\begin{abstract}
Existing algorithms for generating Counterfactual Explanations (CXs) for Machine Learning (ML) typically assume fully specified inputs.
However, real-world data often contains missing values, and the impact of these incomplete inputs on the performance of existing CX methods remains unexplored.
To address this gap, we systematically evaluate recent CX generation methods on their ability to provide valid and plausible counterfactuals when inputs are incomplete. As part of this investigation, we hypothesize that robust CX generation methods will be better suited to address the challenge of providing valid and plausible counterfactuals when inputs are incomplete. Our findings reveal that while robust CX methods achieve higher validity than non-robust ones, all methods struggle to find valid counterfactuals. These results motivate the need for new CX methods capable of handling incomplete inputs.
\end{abstract}

\section{Introduction}Counterfactual Explanations (CXs) describe how the input to a Machine Learning (ML) model should (minimally) change for the model to produce a different, often more desirable outcome. 
Within Explainable AI, CXs have emerged as a leading paradigm for enhancing ML explainability due to their alignment with human reasoning and intelligibility~\cite{Miller_19}. Current explanation algorithms typically require the input, for which a CX is sought, to be fully specified. This can be a limitation in real-world applications, which may involve incomplete or uncertain information. For instance, incomplete inputs may originate from corrupted sensor readings, where some values might be missing, as exemplified below. 

\textbf{\emph{Motivating example.}} \emph{To see what makes missing input values challenging, let us consider the following example based on a real-world application by Tuefekci~\cite{TUFEKCI2014126}, where an ML model $M$ is trained to estimate whether the electrical power $p_e$ produced by a Combined Cycle Power Plant (CCPP) will meet a given target $T$.
In a CCPP, electricity is generated by gas and steam turbines, which are combined in one cycle. 
This process is heavily influenced by four factors: temperature, atmospheric pressure, humidity, and steam pressure. 
These values are collected by sensors placed in the CCPP and are used to train the model $M$. More precisely,  the model $M$ takes these four values as input and produces a binary output equal to $1$ if $p_e \geq T$ and $0$ otherwise. For instance, given an input $x = [18.05, 44.77, 1008.89, 38.78]$, we obtain $M(x) = 0$ for $T = 454W$. In other words, the CCPP does not produce enough power to reach the target under the conditions captured by $x$. CXs can be generated to explain how conditions would have to change for the CCPP to meet its target. For instance, a CX for $x$ would be a new input $x' = [20.57, 44.77, 1008.89, 38.78]$ for which $M(x') = 1$ holds. Now consider a new input $\tilde{x} = [2.5, 30.11, -, 27.35]$ resulting from a faulty sensor reading, where the value for humidity could not be obtained. In this setting, existing CX methods are not applicable to this input due to the missing value. }

\begin{figure}[]
\centering
\begin{tikzpicture}[scale=1.2,>=stealth]

\def\dbpath{(0,3.7)(2,3)(2.5,2.6)(3,2)(3.2,1.8)(3.8,1)(4.5,0.2)}

\fill[magenta!20]
  (0,0)
  -- plot coordinates {\dbpath}
  -- (4.5,0)
  -- cycle;

\fill[green!20]
  plot coordinates {\dbpath}
  -- (4.5,3.7)
  -- (0,3.7)
  -- cycle;

\draw[->] (0,0) -- (4.5,0) node[midway,below]{Steam Pressure};
\draw[->] (0,0) -- (0,3.7) node[midway, rotate=90,yshift=8pt]{Humidity};

\node[circle,fill=black,label=below:$X$]   (Y) at (1,0.8) {};
\node[circle,fill=black,label=below:$X'$]   (Yp) at (3,0.8) {};

\node[circle,fill=black,label=above:$Y$]   (X) at ($(Y)+(0,+2)$) {};
\node[circle,fill=black,label=above:$Y'$]   (Xp) at ($(Yp)+(0,+2)$) {};

\draw[->,thick] (Y) -- node[above]{Recourse $\delta$} (Yp);
\draw[->,thick] (X) -- node[below]{Recourse $\delta$} (Xp);

\draw[dotted,<->] ($(Y)+(0,+0.3)$) -- ($(X)+(0,-0.3)$)
  node[midway,font=\small,xshift=-7pt,rotate=90]{Imputation}
  node[midway,font=\small,xshift=6pt,rotate=90]{Error};
\end{tikzpicture}
\caption{Imputation inaccuracy impedes counterfactual validity~\cite{DBLP:journals/corr/abs-2304-14606}.}
\label{fig:imputation-cx}
\end{figure}
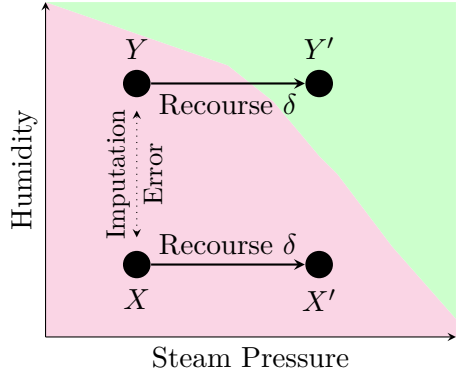

One approach to handling missing data is to employ imputation techniques that replace the missing values with computed estimates. However, this is not a reliable solution for explainability, as CX methods can still fail to produce valid counterfactuals when missing values are approximated through imputation~\cite{DBLP:journals/corr/abs-2304-14606}.
Figure~\ref{fig:imputation-cx} illustrates how inaccurate imputation can invalidate a recourse action undertaken by a user to reach a counterfactual outcome.
In our example in Figure~\ref{fig:imputation-cx}, the true input $X$ represents actual operating conditions in the CCPP, while $Y$ denotes its imputed approximation derived from incomplete sensor readings.
The colors represent the decision boundary of the classifier, with green representing the desired counterfactual outcome.
Computing a recourse vector for the the approximation $Y$ ensures the counterfactual outcome for the approximation $Y$ via $Y' = Y + \delta$.
However, applying the recourse $\delta$ in reality to the actual state $X$ of the CCPP, can result in an invalid counterfactual $X'$ that fails to cross into the desired high-energy region due to imputation error.
As a solution, Kanamori et. al~\cite{DBLP:journals/corr/abs-2304-14606} propose computing counterfactuals that remain valid across multiple imputations to reduce invalidity risk.
However, their work focuses only on discrete values and lacks thorough comparison with recent CX methods.

In this work, we deepen our understanding of how incomplete inputs impact counterfactual validity.
To this end, we first carefully motivate and define the problem. 
From the problem definition, we hypothesize that robust CX generation methods~\cite{JiangL0T24} may be able to mitigate the risk of producing invalid explanations after imputation techniques are applied. 
To validate this hypothesis, we systematically evaluate ten CX generation methods, including both robust and non-robust baselines. 
We show that robust counterfactual methods tend to yield more valid explanations than non-robust ones, although at a higher cost and without ensuring validity in many cases.
Lastly, we dive deeper into the seminal work by Wachter~\cite{Wachter_17} and showcase how the method's hyperparameters impact the validity of counterfactuals.
Our experiments motivate the need for counterfactual frameworks that can reason natively about uncertainty over input features, rather than relying solely on imputation.

\section{Related Work}

Counterfactual explanations for incomplete or uncertain inputs have received limited attention in the literature. 
The only dedicated approach so far is by Kanamori et al.~\cite{DBLP:journals/corr/abs-2304-14606}, who propose generating counterfactuals that remain valid across multiple imputations of missing data. 

Beyond incompleteness, \textit{robustness} has emerged as a central theme in counterfactual research. 
Robust counterfactuals are designed to remain valid despite variations such as model retraining, model multiplicity, or noisy execution~\cite{JiangL0T24}. 
Model retraining can alter decision boundaries after data updates or unlearning; model multiplicity refers to the existence of several near-optimal models during training, requiring counterfactuals that are valid for all models; and noisy execution captures imperfections when recourse recommendations are applied in practice~\cite{JiangL0T24}.
In all these cases, robustness ensures that counterfactual validity persists under uncertainty, making robust methods promising candidates for handling incomplete inputs as well.
In the context of feature attribution, Vo et al.~\cite{DBLP:journals/corr/abs-2407-00411} study how different imputation methods affect the resulting Shapley values of incomplete instances.

Outside of the CX domain, missing data is commonly addressed through imputation techniques that estimate unknown values before downstream processing~\cite{Romero31122025}. 
Selecting an appropriate imputation strategy remains an active area of research in statistics and machine learning, typically depending on data characteristics and evaluation metrics~\cite{doi:10.3233/SJI-230015}. 
To avoid bias toward any particular imputation approach, our empirical analysis considers multiple imputation strategies when evaluating counterfactual generation methods under incomplete inputs.

\newcommand{\pval}{p_{\text{val}}}
\newcommand{\pplaus}{p_{\text{plaus}}}
\newcommand{\pcost}{p_{\text{cost}}}
\newcommand{\pepi}{p_{\text{faith}}}
\newcommand{\cx}{x'}
\newcommand{\rcx}{\delta}
\newcommand{\target}{t}
\section{Problem Definition}
A \textit{counterfactual explanation (CX)} provides insight into a classifier's decision by identifying minimal changes to an input that would lead to a different predicted class~\cite{Wachter_17}.
Formally, let $x \in \mathbb{R}^n$ be a point and $h:\mathbb{R}^n\rightarrow \{0,1\}$ a binary classifier (the extension to multi-class settings is straightforward). Given a distance metric $d: \mathbb{R}^n \times \mathbb{R}^n \rightarrow \mathbb{R}^+$, a CX $\cx \in \mathbb{R}^n$ is a point that minimizes the distance $d(x,\cx)$ subject to $h(\cx) \neq h(x)$. The \emph{cost} of a CX $\cx$ for a point $x$ is defined as $d(x,\cx)$, often measured by an $\ell_p$-norm~\cite{MohammadiKBV21}. Finally, we say that the vector $\delta = \cx - x$ represents the \emph{recourse} action offered by the counterfactual $\cx$. Intuitively, $\delta$ explicitly describes the changes that need to be applied to the original input for the classifier to produce the desired classification.

\textbf{Counterfactuals for Incomplete Inputs}:  
Consider a binary classifier $h$,  
a dataset $D \subset \mathbb{R}^n$ with empirical data distribution $p_D$,  
a distance function $d\colon\mathbb{R}^n \times \mathbb{R}^n \rightarrow \mathbb{R}$,  
and a desired target class label $\target \in \{0,1\}$. Furthermore, let $x \in \mathbb{R}^n$ be a fully specified input and $\tilde{x} \in (\{*\} \cup \mathbb{R})^n$ be an incomplete observation of $x$, such that for each input component $\tilde{x}_i$ with $i \in \{1, \ldots, n\}$, we have either $\tilde{x}_i = x_i$ or $\tilde{x}_i = *$. The goal is to compute a counterfactual $\cx$, and its recourse vector $\rcx$ that satisfies the following criteria:
\begin{description}
    \item[C1] \textbf{Validity of the counterfactual}: The counterfactual $\cx$ is classified as the target class: $h(\cx) = \target$.
    \item[C2] \label{criterion:valid-rc}\textbf{Validity of the recourse}: Applying the recourse $\rcx$ to the original partially observed input $x$ results in classification as the target label: $h(x + \rcx) = \target$.
    \item[C3] \textbf{Cost}: The distance between $\cx$ and $x$ is minimized according to  $d(\cx, x)$. 
\end{description}

Importantly, when computing $\cx$, only the incomplete observation $\tilde{x}$ is available; the fully specified input $x$ remains unknown.
Consequently, the classifier output $h(x)$, from which the intended \emph{counterfactual} outcome $\target = 1 - h(x)$ is typically derived, is also unknown.

In some settings, this limitation renders counterfactuals ill-defined.
For example, in a loan application scenario, a neural network $h$ determines whether a fully specified application $x$ is accepted.
If the application $\tilde{x}$ is incomplete, the prediction $h(\tilde{x})$ is generally undefined.
In this case, a counterfactual explaining how to increase the likelihood of acceptance is not meaningful, since the decision rule itself cannot be evaluated on incomplete inputs, even if the applicant's financial situation improved.

In other settings, however, the classifier output $h(x)$ acts as an \emph{estimator} of an underlying, observable system state that constitutes the factual outcome.
In our running example, if power production in the CCPP becomes too low, an operator may specify a desired high-power state by setting $\target=1$, thereby defining the intended counterfactual outcome.
Here, counterfactuals remain useful despite incomplete observations, as they provide actionable guidance toward a desired system behavior.

Motivated by this distinction, we assume that the desired target class $\target$ is specified by the user.
Accordingly, \textbf{C1} requires only that the counterfactual $\cx$ attains the specified class $\target$.
For \textbf{C2}, recourse validity is evaluated with respect to the true but unseen input $x$, since recourse actions are ultimately applied to the underlying system state, even when it is only partially observed.
Finally, for \textbf{C3}, costs are measured relative to $x$ for the same reason: the cost of recourse depends on the actual system conditions. We illustrate the conceptual difference between \textbf{C1} and \textbf{C2} with an example:\\
\textbf{Example.} Let $h:\mathbb{R}^2\rightarrow\{\top,\bot\}$ be a binary classifier with  $h(x,y):= x+y\geq 4$. Consider an input $x=(1,1)$ that is partially observed via $\tilde{x}=(-,1)$. A point $x' = (2,2)$ is a counterfactual that satisfies \textbf{C1} because $h(x')=2+2\geq 4$. 
From this counterfactual, it is not possible to infer the recourse vector $x'-x$, because $x$ is not known. Using imputation techniques to approximate the missing value in $\tilde{x}$ could yield an inaccurate estimate such as $\hat{x}=(2,1)$.
The recourse vector with respect to $x'$ is then $\delta = (0,1)$ because $\hat{x}+\delta = (2,2)$.
However, this recourse vector does not satisfy \textbf{C2} as it does not flip the classification when applied on the unobserved input: $h(x+\delta) = 1+2 \not\geq 4$.

Notably, recourse validity shares semantic similarities with CX robustness. 
Robustness typically requires that small perturbations to either inputs or model parameters do not change classification outcomes for a counterfactual. 
In contrast, recourse validity requires that uncertainty or incompleteness in some input features does not affect the classification of the counterfactual. 
Building upon this observation, we hypothesize that methods designed to generate robust CXs might inherently offer higher validity. 
Accordingly, in the next section, we design experiments to evaluate whether robust CX methods achieve higher recourse validity compared to non-robust baselines.

\section{Evaluation}

We evaluate ten recent CX generation methods, mostly from the RobustX library~\cite{DBLP:conf/ijcai/JiangMPL25}.  
We exclude only those methods from the RobustX library that are tailored to specialized problems and require auxiliary inputs to ensure a fair comparison.
We also adopt and include ARMIN~\cite{DBLP:journals/corr/abs-2304-14606}, the only method developed specifically for incomplete inputs.
Overall, we compare the following non-robust methods:
\begin{itemize}
    \item BinaryLinearSearch (\texttt{BLS})~\cite{DBLP:conf/aaai/LeofanteP24} finds the closest CX point on the line between the input sample and a randomly selected positive training point.
    \item \texttt{MCE}~\cite{DBLP:conf/aies/MohammadiKBV21} computes the closest point across the decision boundary using mixed integer linear optimization (MILO).
    \item \texttt{Wachter}~\cite{Wachter_17} is similar to \texttt{MCE}, but uses gradient descent instead of MILO.
    \item \texttt{KDTreeNNCE}~\cite{DBLP:journals/datamine/BrughmansLM24} selects the nearest neighbor in the training set from the target class.
    \item \texttt{ARMIN} computes a counterfactual that is valid with respect to multiple imputations of the missing value using MILO. We apply \texttt{ARMIN} to continuous data and use the same margin to the decision threshold as the default value provided by \texttt{MCE}.
\end{itemize}
Furthermore, we compare the following robust methods:
\begin{itemize}
    \item \texttt{MCER}~\cite{JiangL0T23} generates counterfactuals that are robust with respect to a space of delta-perturbations of model weights.
    \item \texttt{RNCE}~\cite{DBLP:journals/ai/JiangLRT24} is similar to \texttt{MCER}, but selects counterfactuals from the training data rather than generating synthetic ones, while maintaining delta-robustness.
    \item \texttt{PROPLACE}~\cite{JiangLL0T23} jointly optimizes plausibility and proximity while preserving robustness.
    \item \texttt{STCE}~\cite{DBLP:conf/icml/DuttaLMTM22,DBLP:conf/icml/HammanNMMD23} generates counterfactuals that are robust with respect to naturally occurring weight changes during retraining.
    \item \texttt{APAS}~\cite{DBLP:conf/ecai/MarzariLCF24} generates robust counterfactuals based on a set of plausible samples representing model changes.
\end{itemize}

To handle incomplete inputs, we apply all three off-the-shelf imputation techniques from the \texttt{scikit-learn} library: Multiple Imputation by Chained Equations (\texttt{MICE})~\cite{JSSv045i03}, k-Nearest Neighbors (\texttt{kNN})~\cite{DBLP:journals/bioinformatics/TroyanskayaCSBHTBA01}, and simple mean imputation (\texttt{Simple}), which replaces missing values with the feature-wise training mean.
Since \texttt{ARMIN} is specifically designed for incomplete inputs, no imputation method is provided by the user.
However, ARMIN internally uses \texttt{MICE} imputation.
To avoid misrepresenting the performance of \texttt{ARMIN}, while still ensuring a fair comparison, we evaluate \texttt{ARMIN} only with \texttt{MICE}, but illuminate how imputation strategies impact the validity of all evaluated methods in Section~\ref{sec:across-dataset}.

We conduct experiments on commonly studied tabular datasets from prior literature: WineQuality~\cite{wine_quality_186}, Diabetes~\cite{Akturk_2020}, Concrete~\cite{YEH19981797}, and Combined Cycle Power Plant~\cite{TUFEKCI2014126}. 
All datasets contain between $4$ and $11$ continuous features, which are min-max normalized. Regression datasets are binarized using a threshold of~$0.5$.
We split each dataset into a training set ($80\%$) and a test set ($20\%)$, and fit a two-layer ReLU-neural network binary classifier $h$ on the training data.
Then, for each test dataset, we randomly sample a batch of $n = 100$ instances $(x_1, \ldots, x_n)$ and record their classifications $c_i := h(x_i)$. Each batch is processed into $9$ experimental setups as follows: For each $m \in \{1, 2, 3\}$, we remove $m$ feature values from every instance completely at random and impute the missing entries using each of the three predefined imputation strategies, resulting in $9$ combinations for each dataset. All counterfactual generation (CX) methods are then applied to these setups to produce counterfactuals $\cx_1, \ldots, \cx_n$ targeting the respective opposite class $t_i := 1-c_i$.
Each resulting set of counterfactuals is evaluated according to four metrics:
\begin{enumerate}
    \item \textbf{Counterfactual Validity (VCX)}: $\sum_{i=1}^n [h(\cx_i) = t_i]/n$ (Measures \textbf{C1}),
    \item \textbf{Recourse Validity (VRC)}: $\sum_{i=1}^n [h(x_i + \rcx_i)= t_i]/n$ (Measures \textbf{C2}),
    \item \textbf{Cost}: $\sum_{i=1}^{n} d(x_i,\cx_i)/n$, where $d$ is the $\ell_1$-distance (Measures \textbf{C3}),
    \item \textbf{Plausibility}: $\sum_{i=1}^n \texttt{lof}(\cx_i)/n$, where \texttt{lof} is the Local Outlier Factor~\cite{DBLP:conf/sigmod/BreunigKNS00}.
\end{enumerate}

The Local Outlier Factor is a metric commonly used to measure plausibility in the context of counterfactual research~\cite{DBLP:conf/icml/DuttaLMTM22,DBLP:conf/acml/JiangLL0T23,DBLP:journals/ai/JiangLRT24}. 
We measure plausibility alongside the main criteria (1-3) to investigate whether plausibility also yields higher validity, an observation reported in prior work in the context of robustness~\cite{PawelczykBK20}.

This diverse experimental setup provides insights that generalize across datasets and imputation strategies, enabling a broad comparison of robust and non-robust CX methods in terms of validity, plausibility, and cost.
To this end, we aggregate each method's scores across the four datasets and three imputation strategies and present the results in Figure~\ref{fig:robust-vs-non-robust-m-1} for one missing value ($m=1$), and in Figure~\ref{fig:robust-vs-non-robust-m-3} for three missing values ($m=3$). 
Results for $m=2$ align with all findings reported here but are omitted to avoid visual overload.

\fboxsep=0pt   
\fboxrule=0.4pt 
\begin{figure}[t]
\centering

\fbox{%
\begin{minipage}[t]{0.48\textwidth}
    \centering
    \resizebox{\textwidth}{!}{\input{robust-vs-non-robust-validity-m=1.pgf}}\\[0.5em]
    \resizebox{\textwidth}{!}{\input{robust-vs-non-robust-cost-m-1.pgf}}\\[0.5em]
    \resizebox{\textwidth}{!}{\input{robust-vs-non-robust-plausibility-m-1.pgf}}\\[0.5em]
    \caption{CX Comparison $m=1$}
    \label{fig:robust-vs-non-robust-m-1}
\end{minipage}
}
\hfill
\fbox{%
\begin{minipage}[t]{0.48\textwidth}
    \centering
    \resizebox{\textwidth}{!}{\input{robust-vs-non-robust-validity-m=3.pgf}}\\[0.5em]
    \resizebox{\textwidth}{!}{\input{robust-vs-non-robust-cost-m-3.pgf}}\\[0.5em]
    \resizebox{\textwidth}{!}{\input{robust-vs-non-robust-plausibility-m-3.pgf}}\\[0.5em]
    \caption{CX Comparison $m=3$}
    \label{fig:robust-vs-non-robust-m-3}
\end{minipage}
}
\end{figure}
\textbf{Recourse Validity (VRC).}  
Figure~\ref{fig:robust-vs-non-robust-m-1}a shows that \textit{robust} methods generally achieve higher validity and exhibit lower variance than \textit{non-robust} methods. 
The higher validity of robust methods over non-robust ones also holds for three missing values ($m=3$), as shown in Figure~\ref{fig:robust-vs-non-robust-m-3}a.
These results confirm that robust methods consistently outperform non-robust approaches in terms of recourse validity, even as the number of missing features increases.
To statistically validate these observations, we compare aggregated VRC scores of robust and non-robust methods using two-sided Mann--Whitney $U$ tests~\cite{497e1044-d5b0-30a9-b230-3ca0f10d6f6c}. 
For $m=1$, robust methods achieve significantly higher recourse validity (median $0.856$ vs.\ $0.713$, $p = 0.000082$). 
And even for $m=3$, the difference remains significant (median $0.703$ vs.\ $0.589$, $p = 0.0026$). 
The only non-robust method that achieves consistently high validity levels comparable to the robust methods is \texttt{ARMIN}, which was specifically designed for incomplete inputs.
Although the low variance of \texttt{ARMIN} is partly due to the limited diversity in the experimental setup, which contains only one imputation method.
The high validity scores of \texttt{ARMIN}, on the other hand, are not merely due to its use of the rather strong imputation method \texttt{MICE}, as we show in the subsequent section, but an actual strength of the method.

However, all methods exhibit shortcomings: First, comparing Figure~\ref{fig:robust-vs-non-robust-m-3}a with Figure~\ref{fig:robust-vs-non-robust-m-1}a reveals that the validity of all methods decreases as the number of missing values increases from one to three. Moreover, even with a single missing value, the validity quantiles of the best-performing methods begin at around $80\%$, indicating relatively low scores for certain experimental configurations.

\textbf{Counterfactual Validity (VCX).} In our experiments, all methods achieve $100\%$ counterfactual validity, except for \texttt{Wachter}, which achieves $84.6\%\pm 14.3\%$. Indeed, Wachter is the only incomplete method, as it relies on gradient-based optimization and can get stuck in local minima. These results are not visualized in any figures due to simplicity.

\textbf{Cost.} Figures~\ref{fig:robust-vs-non-robust-m-1}b and~\ref{fig:robust-vs-non-robust-m-3}b show no clear winner between robust and non-robust methods in terms of cost. 
At first glance, robust methods do not seem to incur additional cost to achieve the aforementioned higher validity compared to non-robust methods. 
However, when comparing the robust methods with the non-robust method \texttt{MCE}, a method that solely optimizes for cost, it is clear that all robust methods incur additional costs -- an insight that was already noted in previous work~\cite{JiangL0T24}. 
Figure~\ref{fig:robust-vs-non-robust-m-1}b and~\ref{fig:robust-vs-non-robust-m-3}b also reveal that \textit{within} robust methods, high validity scores are correlated with high cost, indicating a trade-off between these two objectives. This trade-off, however, is not as apparent within non-robust methods.

\textbf{Plausibility.} Figures~\ref{fig:robust-vs-non-robust-m-1}c and~\ref{fig:robust-vs-non-robust-m-3}c present negative LOF scores for plausibility, where $-1$ marks inliers, and lower values mark out-of-distribution samples.
From these results, we observe that nearly all robust methods produce plausible counterfactuals. This observation aligns with prior findings~\cite{JiangL0T24} that show a link between robustness and plausibility. Furthermore, both \texttt{BLS} and \texttt{KDTreeNNCE} achieve high plausibility scores, likely because they rely on real data points to guide their counterfactual search.
Most notably, \texttt{ARMIN} produces the lowest plausibility scores. 
The additional constraints imposed by \texttt{ARMIN}, which require counterfactual validity across multiple imputations, tend to yield implausible counterfactuals. 
We leave a more detailed investigation of this phenomenon to future work.

\subsection{Across-Dataset Comparison}\label{sec:across-dataset}
While the previous section aggregated the diverse experimental setups into single plots to achieve generalizable insights, this section provides insights into each experimental setup, highlighting the impact of the imputation methods and datasets.
To this end, we show the performance of each method in Table~\ref{tab:combined-data-m-2} for two missing values ($m=2$) for the datasets concrete and power.
We chose these datasets as representatives; the results for the concrete dataset are similar to the results of the Wine and diabetes datasets, whereas the results for the power dataset deviate more and are therefore presented explicitly here.
We note that the results for $m=1$ and $m=3$ align with all findings reported here.  

\textbf{Imputation.}
The $k$-Nearest Neighbors (\texttt{kNN}) imputation method results in higher validity scores at the same cost and same plausibility than the other imputation methods shown in Table~\ref{tab:combined-data-m-2}.
The second best imputation strategy is \texttt{MICE}, which still yields higher validity scores than the trivial \texttt{mean} baseline on all datasets. 
Overall, these results show that the selected imputation strategy plays a vital role for the CX-method's validity.

Given the importance of the imputation strategy, we revisit the performance of \texttt{ARMIN}. Since \texttt{ARMIN} relies exclusively on \texttt{MICE} imputation, we compare it against the other baselines under the same imputation setting in Table~\ref{tab:combined-data-m-2}. 
In this direct comparison, \texttt{ARMIN} achieves higher validity scores than all other baselines on the power dataset, and yields performance comparable to the robust baselines on the concrete dataset. These results indicate that \texttt{ARMIN}'s strong validity cannot be attributed solely to its imputation strategy.

\textbf{Validity.}
Most strikingly, the recourse validity (VRC) of all methods is much lower for the Power dataset than for the Concrete dataset, regardless of whether a method generates robust or non-robust counterfactuals. 
The drop is particularly pronounced for \texttt{Wachter}, which frequently fails to produce even a valid counterfactual (VCX) on the power dataset. 
Indeed, as \texttt{Wachter} relies on gradient descent optimization, it can become trapped in local minima, preventing it from reaching a valid counterfactual.

\textbf{Plausibility.} As expected, all methods that optimize for plausibility achieve consistently plausible counterfactuals on both datasets.
Conversely, the remaining methods, \texttt{MCE, MCER, ARMIN}, and \texttt{Wachter}, produce less plausible counterfactuals for both datasets. 
Interestingly, the plausibility scores of all four non-plausible methods are much lower on the Power dataset than on the Concrete dataset, highlighting that the Power dataset is also challenging with respect to plausibility.

\textbf{Key takeaway:} These findings suggest that both robust and non-robust counterfactual generation methods are sensitive to the characteristics of the underlying data distribution. In practice, this implies that hyperparameters should be carefully tuned for each dataset to ensure reliable validity and plausibility.

\begin{table}[]
\centering
\resizebox{0.65\textwidth}{!}{
\setlength{\tabcolsep}{4pt}

\begin{tabular}{llccccc}
\toprule
 &  & Metric & VRC & VCX & Cost & LOF \\
Method & Dataset & Imputer &  &  &  &  \\
\midrule
\multirow[t]{6}{*}{MCER} & \multirow[t]{3}{*}{Concrete} & KNN & 0.76 & 1.00 & 0.26 $\pm$ 0.20 & -1.12 $\pm$ 0.20 \\
 &  & MICE & 0.64 & 1.00 & 0.36 $\pm$ 0.23 & -1.17 $\pm$ 0.23 \\
 &  & mean & 0.505 & 1.00 & 0.46 $\pm$ 0.22 & -1.25 $\pm$ 0.28 \\
\cline{2-7}
 & \multirow[t]{3}{*}{Power} & KNN & 0.61 & 1.00 & 0.49 $\pm$ 0.20 & -1.87 $\pm$ 0.71 \\
 &  & MICE & 0.53 & 1.00 & 0.52 $\pm$ 0.21 & -1.76 $\pm$ 0.64 \\
 &  & mean & 0.26 & 1.00 & 0.49 $\pm$ 0.21 & -1.46 $\pm$ 0.47 \\
\cline{2-7}
\cline{1-7}
\multirow[t]{6}{*}{PROPLACE} & \multirow[t]{3}{*}{Concrete} & KNN & 0.93 & 1.00 & 0.51 $\pm$ 0.40 & -1.07 $\pm$ 0.08 \\
 &  & MICE & 0.85 & 1.00 & 0.56 $\pm$ 0.41 & -1.10 $\pm$ 0.16 \\
 &  & mean & 0.77 & 1.00 & 0.65 $\pm$ 0.43 & -1.13 $\pm$ 0.19 \\
\cline{2-7}
 & \multirow[t]{3}{*}{Power} & KNN & 0.66 & 1.00 & 0.55 $\pm$ 0.22 & -1.06 $\pm$ 0.07 \\
 &  & MICE & 0.54 & 1.00 & 0.56 $\pm$ 0.24 & -1.06 $\pm$ 0.09 \\
 &  & mean & 0.31 & 1.00 & 0.56 $\pm$ 0.25 & -1.07 $\pm$ 0.13 \\
\cline{2-7}
\cline{1-7}
\multirow[t]{6}{*}{STCE} & \multirow[t]{3}{*}{Concrete} & KNN & 0.95 & 1.00 & 0.70 $\pm$ 0.42 & -1.08 $\pm$ 0.10 \\
 &  & MICE & 0.93 & 1.00 & 0.73 $\pm$ 0.45 & -1.09 $\pm$ 0.14 \\
 &  & mean & 0.86 & 1.00 & 0.83 $\pm$ 0.48 & -1.07 $\pm$ 0.08 \\
\cline{2-7}
 & \multirow[t]{3}{*}{Power} & KNN & 0.77 & 1.00 & 0.61 $\pm$ 0.23 & -1.04 $\pm$ 0.06 \\
 &  & MICE & 0.63 & 1.00 & 0.61 $\pm$ 0.25 & -1.03 $\pm$ 0.07 \\
 &  & mean & 0.39 & 1.00 & 0.62 $\pm$ 0.25 & -1.04 $\pm$ 0.08 \\
\cline{2-7}
\cline{1-7}
\multirow[t]{6}{*}{RNCE} & \multirow[t]{3}{*}{Concrete} & KNN & 0.89 & 1.00 & 0.61 $\pm$ 0.44 & -1.10 $\pm$ 0.12 \\
 &  & MICE & 0.80 & 1.00 & 0.67 $\pm$ 0.46 & -1.09 $\pm$ 0.12 \\
 &  & mean & 0.70 & 1.00 & 0.74 $\pm$ 0.46 & -1.06 $\pm$ 0.08 \\
\cline{2-7}
 & \multirow[t]{3}{*}{Power} & KNN & 0.65 & 1.00 & 0.57 $\pm$ 0.23 & -1.07 $\pm$ 0.07 \\
 &  & MICE & 0.53 & 1.00 & 0.59 $\pm$ 0.23 & -1.06 $\pm$ 0.07 \\
 &  & mean & 0.30 & 1.00 & 0.59 $\pm$ 0.24 & -1.07 $\pm$ 0.16 \\
\cline{2-7}
\cline{1-7}
\multirow[t]{6}{*}{APAS} & \multirow[t]{3}{*}{Concrete} & KNN & 0.93 & 1.00 & 0.64 $\pm$ 0.43 & -1.10 $\pm$ 0.13 \\
 &  & MICE & 0.86 & 1.00 & 0.70 $\pm$ 0.47 & -1.09 $\pm$ 0.12 \\
 &  & mean & 0.78 & 1.00 & 0.76 $\pm$ 0.46 & -1.08 $\pm$ 0.10 \\
\cline{2-7}
 & \multirow[t]{3}{*}{Power} & KNN & 0.71 & 1.00 & 0.59 $\pm$ 0.22 & -1.04 $\pm$ 0.07 \\
 &  & MICE & 0.60 & 1.00 & 0.61 $\pm$ 0.23 & -1.05 $\pm$ 0.08 \\
 &  & mean & 0.35 & 1.00 & 0.60 $\pm$ 0.24 & -1.07 $\pm$ 0.16 \\
\cline{2-7}
\cline{1-7}
\multirow[t]{6}{*}{BLS} & \multirow[t]{3}{*}{Concrete} & KNN & 0.72 & 1.00 & 1.02 $\pm$ 0.57 & -1.17 $\pm$ 0.19 \\
 &  & MICE & 0.66 & 1.00 & 1.03 $\pm$ 0.55 & -1.15 $\pm$ 0.16 \\
 &  & mean & 0.53 & 1.00 & 1.07 $\pm$ 0.57 & -1.20 $\pm$ 0.27 \\
\cline{2-7}
 & \multirow[t]{3}{*}{Power} & KNN & 0.82 & 1.00 & 0.69 $\pm$ 0.28 & -1.09 $\pm$ 0.09 \\
 &  & MICE & 0.62 & 1.00 & 0.69 $\pm$ 0.26 & -1.08 $\pm$ 0.09 \\
 &  & mean & 0.33 & 1.00 & 0.61 $\pm$ 0.25 & -1.14 $\pm$ 0.20 \\
\cline{2-7}
\cline{1-7}
\multirow[t]{6}{*}{KDTreeNNCE} & \multirow[t]{3}{*}{Concrete} & KNN & 0.80 & 1.00 & 0.55 $\pm$ 0.44 & -1.08 $\pm$ 0.12 \\
 &  & MICE & 0.68 & 1.00 & 0.62 $\pm$ 0.47 & -1.08 $\pm$ 0.11 \\
 &  & mean & 0.62 & 1.00 & 0.71 $\pm$ 0.47 & -1.07 $\pm$ 0.10 \\
\cline{2-7}
 & \multirow[t]{3}{*}{Power} & KNN & 0.62 & 1.00 & 0.57 $\pm$ 0.22 & -1.05 $\pm$ 0.07 \\
 &  & MICE & 0.53 & 1.00 & 0.58 $\pm$ 0.22 & -1.05 $\pm$ 0.08 \\
 &  & mean & 0.28 & 1.00 & 0.58 $\pm$ 0.24 & -1.07 $\pm$ 0.16 \\
\cline{2-7}
\cline{1-7}
\multirow[t]{6}{*}{MCE} & \multirow[t]{3}{*}{Concrete} & KNN & 0.505 & 1.00 & 0.24 $\pm$ 0.19 & -1.11 $\pm$ 0.19 \\
 &  & MICE & 0.53 & 1.00 & 0.35 $\pm$ 0.22 & -1.17 $\pm$ 0.22 \\
 &  & mean & 0.44 & 1.00 & 0.44 $\pm$ 0.22 & -1.24 $\pm$ 0.28 \\
\cline{2-7}
 & \multirow[t]{3}{*}{Power} & KNN & 0.505 & 1.00 & 0.49 $\pm$ 0.20 & -1.85 $\pm$ 0.70 \\
 &  & MICE & 0.51 & 1.00 & 0.52 $\pm$ 0.21 & -1.74 $\pm$ 0.63 \\
 &  & mean & 0.23 & 1.00 & 0.49 $\pm$ 0.21 & -1.45 $\pm$ 0.46 \\
\cline{2-7}
\cline{1-7}
\multirow[t]{6}{*}{Wachter} & \multirow[t]{3}{*}{Concrete} & KNN & 0.89 & 1.00 & 0.53 $\pm$ 0.27 & -1.11 $\pm$ 0.15 \\
 &  & MICE & 0.78 & 1.00 & 0.62 $\pm$ 0.26 & -1.20 $\pm$ 0.25 \\
 &  & mean & 0.62 & 1.00 & 0.69 $\pm$ 0.25 & -1.25 $\pm$ 0.27 \\
\cline{2-7}
 & \multirow[t]{3}{*}{Power} & KNN & 0.45 & 0.65 & 0.49 $\pm$ 0.36 & -1.19 $\pm$ 0.30 \\
 &  & MICE & 0.43 & 0.77 & 0.54 $\pm$ 0.31 & -1.18 $\pm$ 0.24 \\
 &  & mean & 0.28 & 0.93 & 0.58 $\pm$ 0.25 & -1.30 $\pm$ 0.37 \\
\cline{2-7}
\cline{1-7}
\multirow[t]{6}{*}{ARMIN} & \multirow[t]{3}{*}{Concrete} & KNN & - & - & - & - \\
 &  & MICE & 0.79 & 1.00 & 0.31 $\pm$ 0.02 & -1.40 $\pm$ 0.03 \\
 &  & mean & - & - & - & - \\
\cline{2-7}
 & \multirow[t]{3}{*}{Power} & KNN & - & - & - & - \\
 &  & MICE & 0.75 & 1.00 & 0.56 $\pm$ 0.01 & -2.87 $\pm$ 0.08 \\
 &  & mean & - & - & - & - \\
\cline{2-7}
\cline{1-7}
\bottomrule
\end{tabular}

}
\caption{Comparison across datasets ($m=2$).}
\label{tab:combined-data-m-2}
\end{table}

\subsection{Hyperparameter Impact of Wachter}
\begin{figure}[]
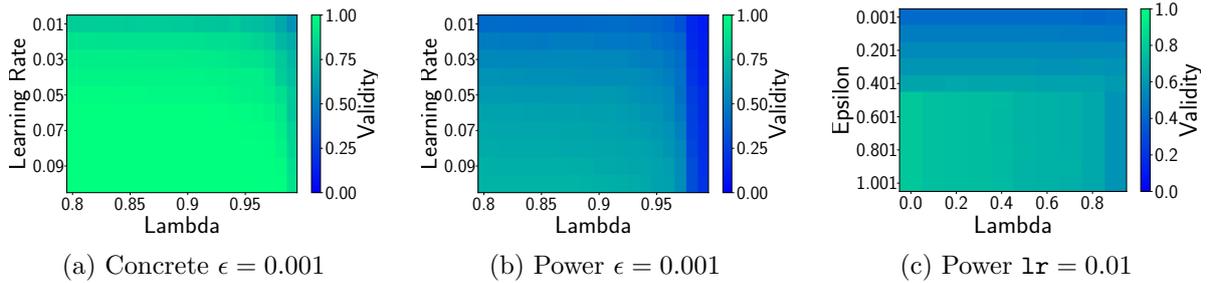

    \centering
    
    \begin{subfigure}[b]{0.32\linewidth}
        \centering
        \resizebox{0.95\textwidth}{!}{\input{heatmap_lr_concrete.pgf}}
        \caption{Concrete $\epsilon = 0.001$}
        \label{fig:wachter-params-concrete-lr}
    \end{subfigure}
    \hfill
    \begin{subfigure}[b]{0.32\linewidth}
        \centering
        \resizebox{0.95\textwidth}{!}{\input{heatmap_lr_power.pgf}}
        \caption{Power $\epsilon = 0.001$}
        \label{fig:wachter-params-power-lr}
    \end{subfigure}
    \hfill
    \begin{subfigure}[b]{0.32\linewidth}
        \centering
        \resizebox{0.95\textwidth}{!}{\input{heatmap_eps_power.pgf}}
        \caption{Power $\texttt{lr}=0.01$}
        \label{fig:wachter-params-power-eps}
    \end{subfigure}
    \caption{Wachter: Hyperparameter Impact on Recourse Validity}
    \label{fig:wachter-params}
\end{figure}

Given the importance of gradient based optimization methods, we focus on the seminal work \texttt{Wachter} and show the influence of its hyperparameters on performance.
We consider three hyperparameters: (a) $\lambda \in [0,1]$, which controls the trade-off between low cost ($\lambda = 1$) and high validity ($\lambda = 0$) in the loss function; (b) $\epsilon > 0$, a stopping criterion based on classifier confidence that terminates the search as soon as the counterfactual has crossed the decision boundary by a margin of $\epsilon$; and (c) the learning rate $\texttt{lr} > 0$, that controls the size of update steps.
Figure~\ref{fig:wachter-params} illustrates the effect of these parameters. Specifically, each axis shows the value of one hyperparameter, and the color visualizes the recourse validity, where bright green colors indicate high validity and dark blue colors indicate low validity. 
The limits for the axes in the plots in Figure~\ref{fig:wachter-params} are selected carefully to visualize only those value ranges that impact validity. 
Each combination of hyperparameter values is evaluated on $100$ test samples from the respective dataset.
For instance, Figure~\ref{fig:wachter-params-concrete-lr} demonstrates that Wachter achieves consistently high validity on the concrete dataset, even for a large trade-off factor ($\lambda = 0.98$) that strongly prioritizes cost over validity. 
The same figure also indicates that increasing the learning rate tends to improve validity. 
We hypothesize that larger learning rates cause more substantial updates during optimization, allowing counterfactuals to cross the decision boundary with a larger step before satisfying the stopping condition.
A similar trend can be observed for the power dataset (Figure~\ref{fig:wachter-params-power-lr}), albeit with overall lower validity, likely due to limitations inherent in Wachter's gradient-based search method.
In both Figures~\ref{fig:wachter-params-concrete-lr} and~\ref{fig:wachter-params-power-lr}, the values of $\lambda$ are  limited to $\lambda \in [0.8,1.0]$ because lower values $\leq 0.8$ did not impact the validity anymore. 
Moreover, as shown in Figure~\ref{fig:wachter-params-power-eps}, adjusting $\epsilon$ also affects recourse validity, yielding improvements comparable to those achieved by increasing the learning rate.
Notably, for high values of $\epsilon \geq 0.5$, the tradeoff parameter $\lambda$ becomes viable for increasing validity.
However, the validity remains low ($80\%$) even when only validity is optimized for and cost is ignored ($\lambda=0$).\\
\textbf{Key takeaway:} While hyperparameter optimization can boost the performance of Wachter, the power dataset remains challenging for this gradient-based optimization method, which can get stuck at local minima.

\section{Conclusion}
We studied the challenge of generating counterfactual explanations (CXs) for incomplete inputs. To this end, we evaluated ten state-of-the-art methods across multiple datasets and imputation strategies and found that robust approaches yield higher validity than non-robust ones. Yet all methods struggle to produce consistently valid explanations. These findings highlight the need for counterfactual frameworks that can reason natively about uncertainty over input features, rather than relying solely on imputation.

\section{Acknowledgements}
This work has been financially supported by Deutsche Forschungsgemeinschaft, DFG Project numbers 459419731 and 434592664.

\begin{footnotesize}

\bibliographystyle{unsrt}   
\bibliography{ctx-missing-input}

\end{footnotesize}
\end{document}